\pdfoutput=1

\documentclass[11pt]{article}

\usepackage[final]{acl}

\usepackage{times}
\usepackage{latexsym}
\usepackage{booktabs}
\usepackage[T1]{fontenc}

\usepackage[utf8]{inputenc}

\usepackage{microtype}

\usepackage{inconsolata}

\usepackage{graphicx}
\usepackage{multirow}
\usepackage{listings}
\usepackage{titlesec}
\titleformat{\paragraph}
{\normalfont\normalsize\bfseries}{\theparagraph}{1em}{}
\titlespacing*{\paragraph}
{0pt}{3.25ex plus 1ex minus .2ex}{1.5ex plus .2ex}

\title{ArgHiTZ at ArchEHR-QA 2025: A Two-Step Divide and Conquer Approach to Patient Question Answering for Top Factuality}

 \author{Adrián Cuadrón\footnotemark[1], Aimar Sagasti\footnotemark[1], Maitane Urruela\thanks{\,This authors contributed equally.}, Iker De la Iglesia, \\ 
 {\bf Ane G Domingo-Aldama, Aitziber Atutxa, Josu Goikoetxea, Ander Barrena }
  \\
 HiTZ Center - Ixa, University of the Basque Country UPV/EHU
 \\
 \texttt{\{adrian.cuadron, aimar.sagasti, maitane.urruela, iker.delaiglesia,}
 \\ \texttt{ane.garciad, aitziber.atutxa, josu.goikoetxea, ander.barrena\}@ehu.eus}
  }

\lstdefinelanguage{json}{
    basicstyle=\ttfamily\footnotesize,
    numbers=left,
    numberstyle=\tiny\color{gray},
    stepnumber=1,
    numbersep=5pt,
    showstringspaces=false,
    breaklines=true, 
    frame=single,
    backgroundcolor=\color{gray!10},
}
\begin{document}
\maketitle

\begin{abstract}
This work presents three different approaches to address the ArchEHR-QA 2025 Shared Task on automated patient question answering. We introduce an end-to-end prompt-based baseline and two two-step methods to divide the task, without utilizing any external knowledge. Both two step approaches first extract \textit{essential} sentences from the clinical text---by prompt or similarity ranking---, and then generate the final answer from these notes. Results indicate that the re-ranker based two-step system performs best, highlighting the importance of selecting the right approach for each subtask. Our best run achieved an overall score of 0.44, ranking 8th out of 30 on the leaderboard, securing the top position in overall factuality. 
\end{abstract}

\section{Introduction}

The increasing volume of patient messages received through online patient portals has become a significant source of clinician burden, highlighting the need for effective automated support. In this context, \textit{ArchEHR-QA 2025: Grounded Electronic Health Record Question Answering Shared Task} \cite{soni-etal-2025-archehr-qa} focuses on the automatic generation of answers to patient-submitted health-related questions, leveraging evidence extracted from their Electronic Health Records (EHRs). The objective of the task is to develop systems capable of producing coherent and evidence-grounded responses, thereby assisting clinicians in managing patient communication more efficiently. In this paper, we describe three different approaches to address the task, present our system design, and analyze its performance on the shared dataset. All the code is publicly available at: \url{https://github.com/hitz-zentroa/ArchEHR-ArgHiTZ}.

\section{Related Work}

Question answering (QA) is a widely used task for evaluating Large Language Models (LLMs), leading to extensive research in both general and domain-specific contexts \cite{yan2024large}. In the medical domain, several benchmark datasets have emerged, like PubMedQA \cite{jin-etal-2019-pubmedqa}, which features yes/no/maybe questions derived from biomedical abstracts, and MedMCQA \cite{pal2022medmcqa}, which contains multiple-choice medical questions. Broader QA tasks, such as Semantic Question Answering (SQA), use datasets like BioASQ-QA \cite{krithara2023bioasq}, for open-domain biomedical questions. More recent work \cite{ben-abacha-etal-2019-overview, ben-abacha-etal-2021-overview} has moved toward patient-centered QA, developing systems that can automatically respond to patient questions using evidence from clinical notes, while evaluating the quality of such responses. The ArchEHR-QA shared task 
builds on this direction by introducing a real-world challenge that aims to respond real patient inquires, relying on their EHRs' information. In this work, we propose and compare multiple strategies to address this task without relying on external medical knowledge, offering insights into effective subtask decomposition for clinical QA.

\section{Data}
In this work, we focus on approaches that require neither additional fine-tuning nor large-scale training, but instead rely on a single example to perform the task. Hence, we exclusively utilize the dataset provided for the task \citep{soni-etal-2025-dataset-patient-needs}, which comprises 20 development (\textit{dev}) documents with patients concerns regarding a specific procedure or treatment they have undergone, along with their corresponding EHRs---derived from MIMIC-III \cite{johnson2016mimic} and MIMIC-IV \cite{johnson2023mimic} databases. These resources are used to generate responses that directly address the patients’ concerns.

Each instance contains the patient's narrative, annotated spans capturing the core question, a technical reformulation, the full EHR, and a sentence-by-sentence breakdown of the EHR with their IDs. A separated file is also provided for the \textit{dev} set with an \textit{essential}, \textit{supplementary} or \textit{not-relevant} relevance classification for each sentence. Our approaches leverage the patient's concern, the technical question, and the classified EHR sentences.

\section{Models} \label{sec:model}

For response generation, we utilized the Aloe \cite{gururajan2024aloe}
, Mistral \cite{mistralai2025mistral7b}
, and Gemma \cite{gemma_2024}
models, as well as, two bigger models for comparison with the task given baseline: Aloe 70B
and Llama 3.3 70B \cite{meta2024llama3.3}
. Additionally, we implemented similarity-based approaches, leveraging different re-ranking models such as Jina \cite{jinaai2024reranker}
, BAAI BGE \cite{li2023making,chen2024bge}
and Alibaba GTE \cite{zhang2024mgte}
.

\section{Evaluation}\label{sec:eval}

The evaluation focuses on the two main objectives of the task: detection of the essential sentences and creation of the response paragraph. To evaluate them, the first step is to check if the responses adhere to the required format: a maximum of 75 words (excluding IDs), and each sentence must end with at least one ID enclosed in vertical bars and separated by comas (e.g., |1| or |2,3|), followed by a line break. If the answer is longer than 75 words, the output is truncated to that limit.

To evaluate the essential sentence detection (\textit{Factuality}) Precision, Recall and F1-Scores are used. The metrics are calculated in two ways: considering only \textit{essential} sentences (\textit{Strict}) and considering both \textit{essential} and \textit{supplementary} sentences (\textit{Lenient}). The generation of the response argument (\textit{Relevance}) is evaluated comparing the generated text with the \textit{essential} sentences and the question using the following metrics: BLEU, ROUGE, SARI, BERTScore, AlignScore and Medcon.

\section{Approaches}
The goal of this shared task is to provide an adequate answer to the patients' health concerns, while citing the source of information---namely, the specific sentence(s) of their EHR. To tackle this challenge, we propose three distinct approaches and systematically compare their performance on the \textit{dev} set in terms of both the quality of the generated answers and the relevance of the cited evidence.

\subsection{Baseline End-to-End} \label{sec:Baseline}

As a baseline, we test various prompts combining \textit{role} prompting---guiding the model to respond as an specific person, in this case a doctor---and Chain of Thought (CoT)---outlining the reasoning steps it must follow before giving the final response---with the models in Section \ref{sec:model} to generate responses in the required format. The final prompt used is shown in Appendix \ref{sec:promptse2e}. Each model receives the patient’s full concern, its reformulation into a technical question, the EHR annotated with sentence IDs, and a one-shot input-output example. We enforce formatting through instructions and light post-processing to fix minor errors.

Table \ref{fig:baseline_ema} shows that the smaller Aloe Beta (8B) model outperforms larger models in both \textit{Factuality} and \textit{Relevance}. Although Gemma achieves the best \textit{Relevance} score, Aloe (8B) remains the overall best option, despite still having room for improvement.
\begin{table}[ht]
\centering
\resizebox{\columnwidth}{!}{%
\begin{tabular}{lccc}
\hline
                 & \multicolumn{3}{c}{Overall Scores}               \\ \hline
Model            & Overall        & Relevance      & Factuality     \\ \hline
\textbf{Aloe 8B} & \textbf{0.388} & 0.312          & \textbf{0.464} \\
Mistral 7B       & 0.364          & 0.327          & 0.402          \\
Gemma 2 9B       & 0.353          & \textbf{0.366} & 0.340          \\
Llama 3.3 70B    & 0.340          & 0.328          & 0.352          \\
Aloe 70B         & 0.371          & 0.332          & 0.410          \\ \hline
\end{tabular}%
}
\caption{Results for the end-to-end with post-process approach in different models. Best results in bold.}
\label{fig:baseline_ema}
\end{table}

Since 70B models show no clear improvement over smaller ones and are computationally costly, we exclude them from other experiments. Manual quality checks also reveal that Gemma, despite its higher \textit{Relevance}, underperforms significantly in \textit{Factuality} and requires more post-processing, so it is also discarded.

\subsection{Two Step Approaches}
Based on the baseline results and limitations, we decide to take two-step approaches in order to split the tasks for the models and reach better performance. On the first step we decide which are the \textit{essential} sentences to respond the patient's concern, by prompting techniques (Section \ref{sec:promt_essentials}) or using a similarity-based re-ranker for the provided sentences (Section \ref{sec:rerank_essentials}). On the second step, those considered sentences are rephrased to build a proper response by prompts and are cited correspondingly following the post-process explained on Section \ref{sec:format}.

\subsubsection{First Step}

\paragraph{Prompting to Extract Essentials} \label{sec:promt_essentials}

In the first approach to identify essential sentences, we leverage a large language model (LLM) using two different techniques, experimenting with several prompting strategies.

\textbf{Extract list of essential sentences:} This technique consists of prompting the model to generate a list of IDs of the essential sentences given the list of all the sentences of the clinical note. This method leverages the model's ability to comprehend the clinical question and utilize the sentences as contextual information to produce an accurate list of IDs. As the input, we use the patient narrative, the clinician question and the clinical note sentences.

\textbf{Determine essentials individually:} In this method, we evaluate each sentence individually to determine whether it should be classified as essential or not. The prompt instructs the model to determine if the information contained in a clinical note sentence is essential for accurately answering the patient's question. To simplify the task, the model is required to respond with a binary "Yes" or "No", reducing the complexity of the output and potentially improving reliability.

We initially employed a basic prompt for each method, which yielded suboptimal results. To improve response accuracy, we subsequently introduced a role-based prompt design by assigning the model the role of a medical expert. Additionally, we incorporated a CoT prompting strategy and one-shot and few-shot prompting techniques, providing the model with concrete examples to guide its responses. An overview of all prompt configurations is provided in Appendices \ref{sec:promptslist} and \ref{sec:promptsindiv}.

To evaluate the quality of the generated lists, we focused solely on the strict metric, prioritizing the F1 score. The results of these two techniques are shown in Table \ref{tab:prompt-essentials} for extracting the list directly and Table \ref{tab:prompt-essentials-individual} for individual technique.

\begin{table}[ht]
\centering
\resizebox{\columnwidth}{!}{%
\begin{tabular}{cllll}
\hline
\textbf{}                      & \multicolumn{1}{c}{\textbf{}} & \multicolumn{3}{c}{\textbf{Strict}}           \\ \hline
\multicolumn{1}{l}{Model}      & Prompt type                   & Prec.         & Rec.          & F1            \\ \hline
\multirow{4}{*}{Aloe 8B}       & basic                         & 0.40          & 0.26          & 0.31          \\
                               & role-based                    & 0.52 & 0.49 & \textbf{0.50} \\
                               & + CoT                    & 0.47          & 0.38          & 0.42          \\
                               & + one-shot         & 0.48          & 0.45          & 0.46          \\
\multicolumn{1}{l}{Aloe 70B}   & role-based                    & 0.52          & 0.30          & 0.38          \\
\multicolumn{1}{l}{Mistral 7B} & role-based                    & 0.55          & 0.42          & 0.48          \\ \hline
\end{tabular}%
}
\caption{Results across models and prompt types to extract \textit{essential} notes for the first step, using a prompt to generate the \textit{essential} lists directly.}
\label{tab:prompt-essentials}
\end{table}

\begin{table}[ht]
\resizebox{\columnwidth}{!}{%
\begin{tabular}{cllll}
\hline
\textbf{}                    & \multicolumn{1}{c}{\textbf{}} & \multicolumn{3}{c}{\textbf{Strict}}           \\ \hline
\multicolumn{1}{l}{Model}    & Prompt type                   & Prec.         & Rec.          & F1            \\ \hline
\multirow{4}{*}{Aloe 8B}     & basic                         & 0.32          & 0.35          & 0.33          \\
                             & role-based                    & 0.38 & 0.58 & \textbf{0.46} \\
                             & + CoT                    & 0.33          & 0.62          & 0.44          \\
                             & + few-shot         & 0.35          & 0.43          & 0.39          \\
\multicolumn{1}{l}{Aloe 70B} & role-based                    & 0.52          & 0.19          & 0.28          \\ \hline
\end{tabular}%
}
\caption{Results across models and prompt types to extract \textit{essential} notes individually in the first step.}
\label{tab:prompt-essentials-individual}
\end{table}

\paragraph{Re-ranker to Extract Essentials}\label{sec:rerank_essentials}
Our second approach is inspired by the typical RAG architecture, using a similarity based ranking model to identify which of the retrieved text chunks---the clinical notes' sentences---are more relevant given a query---a combination of the patient's narrative and the clinician's question. We leverage this method as it aligns with the task's goal of identifying the sentences of the clinical text that are relevant to answer the patient's query.

To determine sentence relevance, we rely on the output scores provided by the reranker and establish a threshold. Sentences with scores above it are labeled as \textit{essential}, while those below are considered \textit{not-relevant}. The optimal threshold is determined by computing the ROC curve and selecting the point that maximizes the Youden index, which allows us to identify the threshold that provides the best trade-off between true positive and false positive rates. However, a potential limitation of this method is that the threshold is determined based on the \textit{dev} set and may not generalize well to the \textit{test} set if the data distribution is different.

\begin{table}[ht]
\centering
\resizebox{\columnwidth}{!}{%
\begin{tabular}{ccccccc}
\hline
 & \multicolumn{3}{c}{Strict} & \multicolumn{3}{c}{Lenient} \\ \hline
Model & Prec. & Rec. & F1 & Prec. & Rec. & F1 \\ \hline
\textbf{Jina} & 0.427 & 0.717 & \textbf{0.535} & 0.547 & 0.672 & \textbf{0.603} \\
Alibaba & 0.422 & 0.681 & 0.521 & 0.552 & 0.651 & 0.597 \\
BAAI & 0.507 & 0.507 & 0.507 & 0.587 & 0.429 & 0.495 \\ \hline
\end{tabular}
}
\caption{ Precision, Recall, and F1-score for each reranking model in the task of identifying essential sentences.}
\label{tab:reranker-models}
\end{table}

\begin{table*}[!ht]
\centering
\resizebox{\textwidth}{!}{%
\begin{tabular}{clccccccc}
\hline
       &     & \multicolumn{3}{c}{Overall Scores}& \multicolumn{2}{c}{Strict}      & \multicolumn{2}{c}{Lenient}     \\ \hline
Data   & Approach            & Overall        & Relevance      & Factuality     & Macro-F1       & Micro-F1       & Macro-F1       & Micro-F1       \\ \hline
\multirow{3}{*}{Dev}  & End-to-End            & 0.388          & \textbf{0.312} & 0.464          & 0.464          & 0.464          & 0.550          & 0.529          \\
       & Two-Step Prompting     & 0.385          & 0.265          & 0.504          & 0.518          & 0.504          & 0.547          & 0.511          \\
       & Two-Step Re-Ranker& \textbf{0.421} & 0.285          & \textbf{0.558} & \textbf{0.566} & \textbf{0.558} & \textbf{0.645} & \textbf{0.626} \\ \hline
\multirow{3}{*}{Test} & End-to-End   &   0.367   &  \textbf{0.325}  &  0.408  &  0.451 &  0.408  &   0.463   &  0.418  \\
       & Two-Step Prompting          &   0.366  &  0.281  &  0.452  &  0.474  &  0.452  &   0.493   &  0.458  \\
       & Two-Step Re-Ranker      & \textbf{0.440} & 0.276          & \textbf{0.605} & \textbf{0.585} & \textbf{0.605} & \textbf{0.619} & \textbf{0.621} \\ \hline
\end{tabular}
}
\caption{Best results of \textit{Relevance} and \textit{Factuality} of the three approaches in \textit{dev} and test sets: Section \ref{sec:Baseline} approach using Aloe (8B), two-step prompting approach using Aloe (8B) in both steps (Section \ref{sec:promt_essentials}) and two-step re-ranker approach leveraging Jina and Aloe (8B) for each step (Section \ref{sec:rerank_essentials}). Best results in bold.}
\label{tab:dev_results}%
\end{table*}

Table \ref{tab:reranker-models} shows the results obtained by different reranker models in predicting \textit{essential} sentences on the \textit{dev} set. As shown in the table, the different models achieve similar F1-scores, with the Jina model obtaining the highest F1-score among them.

\subsubsection{Second Step} \label{sec:format}

After extracting essential sentences, we prompt a model to generate a response addressing the patient's concerns using only those sentences, without citations. Following the strategy used in Section \ref{sec:Baseline}, one example is shown, but based solely on the extracted essentials rather than the full EHR.

The generated response is then post-processed to (1) meet a 75-word limit and (2) add citations. For the first, the given response is split into sentences, and we do a selection to stay within the word limit. For the second, we match them to their most similar \textit{essential} sentence of the first step, based on similarity scores. Citations are added accordingly, capping the number per sentence for balance. Title-like sentences from the first step are excluded in this step. Therefore, this process preserves---and often improves---the F-Score of the first step, while ensuring a coherent, well-cited response. We leverage Aloe (8B) to get the response since it seems to be the best model in previous steps. Table \ref{tab:dev_results} shows the results after performing the second step.

\section{Discussion}
After obtaining the \textit{dev} and \textit{test} results (Table \ref{tab:dev_results}), it becomes evident that the best-performing strategy is the two-step approach with a reranker for \textit{essential} sentence extraction, while the prompt-based two-step variant does not improve end-to-end overall results.

This suggests that more critical than merely dividing the task into smaller and simpler subtasks, is the choice of an appropriate method for each subtask. In this case, using a reranker to extract the most relevant sentences of the EHR for the patient's question proves to be more effective than the prompting-based selection. 

Additionally, all three of our approaches surpass the organizers' zero-shot baseline, which achieves overall scores of 0.359 and 0.307 on the \textit{dev} and \textit{test} sets, respectively, despite relying on a significantly larger model (Llama 3.3 70B). This further supports our conclusion in Section \ref{sec:Baseline} that larger models do not necessarily outperform smaller ones in complex tasks requiring not only medical expertise but also argumentative, summarization, and rewriting skills. Notably, our best run achieves an overall score of \textbf{0.44}, ranking \textbf{8\textsuperscript{th} out of 30} on the test leaderboard and obtaining the top position in overall \textit{Factuality}, even though it does not rely on external knowledge.
Nonetheless, our three systems---particularly the two-step ones---exhibit certain limitations in \textit{Relevance}, suggesting that alternative methods could be explored in future work to improve the response drafting process.

\section{Conclusions}
This work presents three approaches to respond to patient inbox messages using only the patient's concern, its technical reformulation, and the corresponding EHR. Our methods accurately generate responses and identify relevant information sources in the given text without relying on external data. Furthermore, we find that splitting the task into smaller, targeted subtasks improves performance when each is addressed with tailored methods. Future work may explore alternative response formulations to enhance clarity and improve the relevance score. In conclusion, we demonstrate that accurate message response is achievable without training data or external information.

\section*{Acknowledgments}
This work has been partially supported by the HiTZ Center and the Basque Government (IXA excellence research group funding IT-1570-22 and IKER-GAITU project), as well as by the Spanish Ministry of Universities, Science and Innovation MCIN/AEI/10.13039/501100011033 by means of the projects: Proyectos de Generación de Conocimiento 2022 (EDHER-MED/EDHIA PID2022-136522OB-C22), DeepKnowledge (PID2021-127777OB-C21) and DeepMinor (CNS2023-144375). It is also supported by ILENIA (2022/TL22/00215335) and EU NextGeneration EU/PRTR (DeepR3 TED2021-130295B-C31) projects. And also by an FPU grant (Formación de Profesorado Universitario) from the Spanish Ministry of Science, Innovation and Universities (MCIU) to the fourth author (FPU23/03347).

\section*{Limitations}
In this work we leverage several Instruct models, as well as, one and few-shot prompting techniques. Due to the lack of training samples, we do not extend the prompt examples and neither perform any finetuning. Additionally, in the two-step systems, there is still room for improvement in order to enhance the \textit{Relevance} overall score---for instance, by trying text-to-text models like T5 \cite{2020t5}. In this work we focus on techniques that utilize only the available data, therefore Information Retrieval methods such as Retrieval Augmented Generation (RAG) are not employed, even though they could be beneficial given the limited data available. We leave these methods for future work.

\bibliography{acl_latex}

\begin{thebibliography}{19}
\providecommand{\natexlab}[1]{#1}

\bibitem[{Ben~Abacha et~al.(2021)Ben~Abacha, Mrabet, Zhang, Shivade, Langlotz, and Demner-Fushman}]{ben-abacha-etal-2021-overview}
Asma Ben~Abacha, Yassine Mrabet, Yuhao Zhang, Chaitanya Shivade, Curtis Langlotz, and Dina Demner-Fushman. 2021.
\newblock \href {https://doi.org/10.18653/v1/2021.bionlp-1.8} {Overview of the {MEDIQA} 2021 shared task on summarization in the medical domain}.
\newblock In \emph{Proceedings of the 20th Workshop on Biomedical Language Processing}, pages 74--85, Online. Association for Computational Linguistics.

\bibitem[{Ben~Abacha et~al.(2019)Ben~Abacha, Shivade, and Demner-Fushman}]{ben-abacha-etal-2019-overview}
Asma Ben~Abacha, Chaitanya Shivade, and Dina Demner-Fushman. 2019.
\newblock \href {https://doi.org/10.18653/v1/W19-5039} {Overview of the {MEDIQA} 2019 shared task on textual inference, question entailment and question answering}.
\newblock In \emph{Proceedings of the 18th BioNLP Workshop and Shared Task}, pages 370--379, Florence, Italy. Association for Computational Linguistics.

\bibitem[{Chen et~al.(2024)Chen, Xiao, Zhang, Luo, Lian, and Liu}]{chen2024bge}
Jianlv Chen, Shitao Xiao, Peitian Zhang, Kun Luo, Defu Lian, and Zheng Liu. 2024.
\newblock \href {https://arxiv.org/abs/2402.03216} {Bge m3-embedding: Multi-lingual, multi-functionality, multi-granularity text embeddings through self-knowledge distillation}.
\newblock \emph{Preprint}, arXiv:2402.03216.

\bibitem[{Gururajan et~al.(2024)Gururajan, Lopez-Cuena, Bayarri-Planas, Tormos, Hinjos, Bernabeu-Perez, Arias-Duart, Martin-Torres, Urcelay-Ganzabal, Gonzalez-Mallo, Alvarez-Napagao, Ayguadé-Parra, and Garcia-Gasulla}]{gururajan2024aloe}
Ashwin~Kumar Gururajan, Enrique Lopez-Cuena, Jordi Bayarri-Planas, Adrian Tormos, Daniel Hinjos, Pablo Bernabeu-Perez, Anna Arias-Duart, Pablo~Agustin Martin-Torres, Lucia Urcelay-Ganzabal, Marta Gonzalez-Mallo, Sergio Alvarez-Napagao, Eduard Ayguadé-Parra, and Ulises Cortés~Dario Garcia-Gasulla. 2024.
\newblock \href {https://arxiv.org/abs/2405.01886} {Aloe: A family of fine-tuned open healthcare llms}.
\newblock \emph{Preprint}, arXiv:2405.01886.

\bibitem[{Jin et~al.(2019)Jin, Dhingra, Liu, Cohen, and Lu}]{jin-etal-2019-pubmedqa}
Qiao Jin, Bhuwan Dhingra, Zhengping Liu, William Cohen, and Xinghua Lu. 2019.
\newblock \href {https://doi.org/10.18653/v1/D19-1259} {{P}ub{M}ed{QA}: A dataset for biomedical research question answering}.
\newblock In \emph{Proceedings of the 2019 Conference on Empirical Methods in Natural Language Processing and the 9th International Joint Conference on Natural Language Processing (EMNLP-IJCNLP)}, pages 2567--2577, Hong Kong, China. Association for Computational Linguistics.

\bibitem[{JinaAI(2024)}]{jinaai2024reranker}
JinaAI. 2024.
\newblock jina-reranker-v2-base-multilingual.
\newblock \url{https://hf.co/jinaai/jina-reranker-v2-base-multilingual}.

\bibitem[{Johnson et~al.(2023)Johnson, Pollard, Horng, Celi, and Mark}]{johnson2023mimic}
Alistair Johnson, Tom Pollard, Steven Horng, Leo~Anthony Celi, and Roger Mark. 2023.
\newblock Mimic-iv-note: Deidentified free-text clinical notes (version 2.2). physionet.

\bibitem[{Johnson et~al.(2016)Johnson, Pollard, Shen, Lehman, Feng, Ghassemi, Moody, Szolovits, Anthony~Celi, and Mark}]{johnson2016mimic}
Alistair~EW Johnson, Tom~J Pollard, Lu~Shen, Li-wei~H Lehman, Mengling Feng, Mohammad Ghassemi, Benjamin Moody, Peter Szolovits, Leo Anthony~Celi, and Roger~G Mark. 2016.
\newblock Mimic-iii, a freely accessible critical care database.
\newblock \emph{Scientific data}, 3(1):1--9.

\bibitem[{Krithara et~al.(2023)Krithara, Nentidis, Bougiatiotis, and Paliouras}]{krithara2023bioasq}
Anastasia Krithara, Anastasios Nentidis, Konstantinos Bougiatiotis, and Georgios Paliouras. 2023.
\newblock Bioasq-qa: A manually curated corpus for biomedical question answering.
\newblock \emph{Scientific Data}, 10(1):170.

\bibitem[{Li et~al.(2023)Li, Liu, Xiao, and Shao}]{li2023making}
Chaofan Li, Zheng Liu, Shitao Xiao, and Yingxia Shao. 2023.
\newblock \href {https://arxiv.org/abs/2312.15503} {Making large language models a better foundation for dense retrieval}.
\newblock \emph{Preprint}, arXiv:2312.15503.

\bibitem[{MetaAI(2024)}]{meta2024llama3.3}
MetaAI. 2024.
\newblock Llama-3.3-70b-instruct.
\newblock \url{https://hf.co/meta-llama/Llama-3.3-70B-Instruct}.

\bibitem[{MistralAI(2025)}]{mistralai2025mistral7b}
MistralAI. 2025.
\newblock Mistral-7b-instruct-v0.3.
\newblock \url{https://hf.co/mistralai/Mistral-7B-Instruct-v0.3}.

\bibitem[{Pal et~al.(2022)Pal, Umapathi, and Sankarasubbu}]{pal2022medmcqa}
Ankit Pal, Logesh~Kumar Umapathi, and Malaikannan Sankarasubbu. 2022.
\newblock Medmcqa: A large-scale multi-subject multi-choice dataset for medical domain question answering.
\newblock In \emph{Conference on health, inference, and learning}, pages 248--260. PMLR.

\bibitem[{Raffel et~al.(2020)Raffel, Shazeer, Roberts, Lee, Narang, Matena, Zhou, Li, and Liu}]{2020t5}
Colin Raffel, Noam Shazeer, Adam Roberts, Katherine Lee, Sharan Narang, Michael Matena, Yanqi Zhou, Wei Li, and Peter~J. Liu. 2020.
\newblock \href {http://jmlr.org/papers/v21/20-074.html} {Exploring the limits of transfer learning with a unified text-to-text transformer}.
\newblock \emph{Journal of Machine Learning Research}, 21(140):1--67.

\bibitem[{Soni and Demner-Fushman(2025{\natexlab{a}})}]{soni-etal-2025-dataset-patient-needs}
Sarvesh Soni and Dina Demner-Fushman. 2025{\natexlab{a}}.
\newblock A dataset for addressing patient's information needs related to clinical course of hospitalization.
\newblock \emph{arXiv preprint}.

\bibitem[{Soni and Demner-Fushman(2025{\natexlab{b}})}]{soni-etal-2025-archehr-qa}
Sarvesh Soni and Dina Demner-Fushman. 2025{\natexlab{b}}.
\newblock Overview of the archehr-qa 2025 shared task on grounded question answering from electronic health records.
\newblock In \emph{The 24th Workshop on Biomedical Natural Language Processing and BioNLP Shared Tasks}, Vienna, Austria. Association for Computational Linguistics.

\bibitem[{Team(2024)}]{gemma_2024}
Gemma Team. 2024.
\newblock \href {https://doi.org/10.34740/KAGGLE/M/3301} {Gemma}.

\bibitem[{Yan et~al.(2024)Yan, Niu, Li, Zhang, Yin, Fei, Peng, Bi, Feng, Chen et~al.}]{yan2024large}
Lawrence~KQ Yan, Qian Niu, Ming Li, Yichao Zhang, Caitlyn~Heqi Yin, Cheng Fei, Benji Peng, Ziqian Bi, Pohsun Feng, Keyu Chen, and 1 others. 2024.
\newblock Large language model benchmarks in medical tasks.
\newblock \emph{arXiv preprint arXiv:2410.21348}.

\bibitem[{Zhang et~al.(2024)Zhang, Zhang, Long, Xie, Dai, Tang, Lin, Yang, Xie, Huang et~al.}]{zhang2024mgte}
Xin Zhang, Yanzhao Zhang, Dingkun Long, Wen Xie, Ziqi Dai, Jialong Tang, Huan Lin, Baosong Yang, Pengjun Xie, Fei Huang, and 1 others. 2024.
\newblock mgte: Generalized long-context text representation and reranking models for multilingual text retrieval.
\newblock In \emph{Proceedings of the 2024 Conference on Empirical Methods in Natural Language Processing: Industry Track}, pages 1393--1412.

\end{thebibliography}
\UseRawInputEncoding

\newpage
\appendix
\onecolumn
\section{Prompts}
\label{sec:prompts}
\subsection{End-to-End Prompt}
\label{sec:promptse2e}
\begin{lstlisting}[language=json, breaklines=true]
{
    "system": "You are a Medical Report Assistant. Your role is to generate a coherent paragraph that answers a patient's question using only the information you consider relevant from the patient's medical record. You will receive as input:\n1. A narrative of the patient.\n2. The patient's question reformulated by a doctor.\n3. The medical record with the necessary information to answer the patient concern.\n\nYour task is to produce a paragraph answer that:\n- Rephrases and integrates the information from the provided notes without copying them verbatim.\n- Selects only the sentences you deem relevant to answer the question (do not use sentences that do not add value to the answer).\n- Clearly cites the sentence numbers that contributed to each sentence in your response, with the citation placed immediately after the sentence enclosed in vertical bars (e.g., |1| or |2,3|).\n- Does not repeat or omit any sentence that you consider relevant, and does not invent any additional sentences.\n- Contains a maximum of 75 words in total.\n- Strictly adheres to the following format: each sentence on a new line with its citation; no additional text or explanations.{example_case}\n\nEnsure your final output strictly follows this format: one sentence per line with its corresponding citation, using the provided sentences you consider relevant (without any repetition or addition), and the total output does not exceed 75 words.\n\n- Ensure clarity, brevity, and accuracy in your response. Here goes an example:\n\nExample Input:\nCase: 0\nPatient Narrative:\nTook my 59 yo father to ER ultrasound discovered he had an aortic aneurysm. He had a salvage repair (tube graft). Long surgery / recovery for couple hours then removed packs. why did they do this surgery????? After this time he spent 1 month in hospital now sent home.\n\nReformulated Question:\nWhy did they perform the emergency salvage repair on him?\n\nClinical Notes:\n1: He was transferred to the hospital on 2025-1-20 for emergent repair of his ruptured thoracoabdominal aortic aneurysm. 2: He was immediately taken to the operating room where he underwent an emergent salvage repair of ruptured thoracoabdominal aortic aneurysm with a 34-mm Dacron tube graft using deep hypothermic circulatory arrest. 3: Please see operative note for details which included cardiac arrest x2. 4: Postoperatively he was taken to the intensive care unit for monitoring with an open chest. 5: He remained intubated and sedated on pressors and inotropes. 6: On 2025-1-22, he returned to the operating room where he underwent exploration and chest closure. 7: On 1-25 he returned to the OR for abd closure JP/ drain placement/ feeding jejunostomy placed at that time for nutritional support.\n\n8: Thoracoabdominal wound healing well with exception of very small open area mid wound that is @1cm around and 1/2cm deep, no surrounding erythema. 9: Packed with dry gauze and covered w/DSD.
    
    \n\nAnswer:\nHis aortic aneurysm was caused by the rupture of a thoracoabdominal aortic aneurysm, which required emergent surgical intervention. |1|\n He underwent a complex salvage repair using a 34-mm Dacron tube graft and deep hypothermic circulatory arrest to address the rupture. |2|\n The extended recovery time and hospital stay were necessary due to the severity of the rupture and the complexity of the surgery, though his wound is now healing well with only a small open area noted. |8|\n\n Now the REAL CASE:\n",
  "user": "Answer to the patient using the following inputs: Case: {id}\n\nPatient Narrative: {patient_narrative}, \n Reformulated Question: {clinician_question}\n  Clinical Notes: \n{note_excerpt} \n\nNow write the output paragraph based solely on the sentences you consider relevant to answer the question. Your output must:\n- Use only the sentences that you consider relevant from the provided text without repeating any sentence.\n- Contain a maximum of 75 words.\n- Include the citation (sentence number(s) used) immediately after each sentence, within vertical bars.\n- Follow exactly the format described: one sentence per line with its citation, and no additional text. Your final output must be the answer paragraph only, with no extra explanation or text.",
  "assistant": "Answer:"
}
\end{lstlisting}
\newpage
\subsection{Prompts to Extract List of Essentials}
\label{sec:promptslist}
\subsubsection{Role-Based Prompt}
\begin{lstlisting}[language=json, breaklines=true]
{
    "system": "You are a medical expert.\nYou are given:\n\nA patient narrative written by a family member or caregiver.\n\nA clinical question derived from the narrative.\n\nA list of numbered clinical notes from the patient's medical record.\n\nYour task is to identify which of the clinical notes are essential for answering the clinical question.\nReturn only the numbers of the essential notes in a comma-separated list.\nDo not explain your reasoning. Just return the list.",
    "user": "Patient Narrative:\n {patient_narrative}\n\nClinical Question:\n {clinical_question} \n\nClinical Notes:\n {sentences}"
}
\end{lstlisting}

\subsubsection{CoT Prompt}
\begin{lstlisting}[language=json, breaklines=true]
{
    "system": "You are a medical expert with extensive experience in clinical natural language processing, specializing in extracting key information from clinical notes to answer medical questions. Your deep clinical knowledge and expertise in the healthcare domain enable you to identify critical data points from complex medical texts.\n\nTask: You will be provided with the patient narrative and the clinical question, and a set of clinical notes (each sentence is assigned a unique ID). Your goal is to identify only the sentences that contain critical information needed to answer the clinical question.\n\nInstructions:\n1. Internally, perform a detailed step-by-step analysis (chain-of-thought) of the clinical question and each clinical note. Evaluate each sentence for key information, context, and relevance.\n2. Select only the sentences that contain essential information to answer the question.\n3. Return only the IDs of those sentences, without including any additional text or explanation.\n\nOutput Format:\nA list of the relevant sentence IDs, separated by commas.\n\nReminder: Use your internal chain-of-thought to reason through the task, but do not display any of that reasoning in your final output. Simply provide the final answer as the list of IDs.",
    "user": "Patient Narrative:\n {patient_narrative}\n\nQuestion:\n {clinical_question} \n\nClinical Notes:\n {sentences}"
}
\end{lstlisting}

\subsubsection{CoT + One-Shot Prompt}
\begin{lstlisting}[language=json, breaklines=true]
{
    "system": "You are a medical expert with extensive experience in clinical natural language processing, specializing in extracting key information from clinical notes to answer medical questions. Your deep clinical knowledge and expertise in the healthcare domain enable you to identify critical data points from complex medical texts.\n\nTask: You will be provided with the patient narrative and the clinical question, and a set of clinical notes (each sentence is assigned a unique ID). Your goal is to identify only the sentences that contain critical information needed to answer the clinical question.\n\nInstructions:\n1. Internally, perform a detailed step-by-step analysis (chain-of-thought) of the clinical question and each clinical note. Evaluate each sentence for key information, context, and relevance.\n2. Select only the sentences that contain essential information to answer the question.\n3. Return only the IDs of those sentences, without including any additional text or explanation.\n\nOutput Format:\nA list of the relevant sentence IDs, separated by commas.\n\nReminder: Use your internal chain-of-thought to reason through the task, but do not display any of that reasoning in your final output. Simply provide the final answer as the list of IDs.\n\nHere is an example of the output:\n\nPatient Narrative:\nTook my 59 yo father to ER ultrasound discovered he had an aortic aneurysm. He had a salvage repair (tube graft). Long surgery / recovery for couple hours then removed packs. why did they do this surgery????? After this time he spent 1 month in hospital now sent home.\n\nClinical Question:\nWhy did they perform the emergency salvage repair on him?\n\nCinical notes: \n1: He was transferred to the hospital on 2025-1-20 for emergent repair of his ruptured thoracoabdominal aortic aneurysm. 2: He was immediately taken to the operating room where he underwent an emergent salvage repair of ruptured thoracoabdominal aortic aneurysm with a 34-mm Dacron tube graft using deep hypothermic circulatory arrest. 3: Please see operative note for details which included cardiac arrest x2. 4: Postoperatively he was taken to the intensive care unit for monitoring with an open chest. 5: He remained intubated and sedated on pressors and inotropes. 6: On 2025-1-22, he returned to the operating room where he underwent exploration and chest closure. 7: On 1-25 he returned to the OR for abd closure JP/ drain placement/ feeding jejunostomy placed at that time for nutritional support. 8: Thoracoabdominal wound healing well with exception of very small open area mid wound that is @1cm around and 1/2cm deep, no surrounding erythema. 9: Packed with dry gauze and covered w/DSD.\n\nAnswer:\nList of essential clinical notes::: 1, 2, 8",
    "user": "Patient Narrative:\n {patient_narrative}\n\nClinical Question:\n {clinical_question} \n\nClinical Notes:\n {sentences}"
}
\end{lstlisting}

\subsection{Prompts to Extract Essentials Individually}
\label{sec:promptsindiv}
\subsubsection{Role-Based Prompt}
\begin{lstlisting}[language=json, breaklines=true]
{
    "system":"You are a medical expert.\nYou are given:\n\nA patient narrative written by a family member or caregiver.\n\nA clinical question derived from the narrative.\n\nA single clinical note extracted from the patient’s medical record.\n\nYour task is to determine whether the information in the clinical note is essential to accurately answer the clinical question.\nRespond with only one word: Yes or No.",
    "user": "Patient Narrative:\n {patient_narrative}\n\nClinical Question:\n {clinical_question}\n\nClinical Note:\n {sentences}"
}
\end{lstlisting}

\subsubsection{CoT Prompt}
\begin{lstlisting}[language=json]
{
    "system": "You are a medical expert with extensive experience in clinical natural language processing, specializing in analyzing clinical notes to assess patient conditions. Your deep clinical knowledge enables you to accurately interpret and evaluate medical narratives.\n\nTask: You will be provided with the patient narrative, the clinical question and a clinical note of the patient's clinical history. Your goal is to determine whether the clinical note contains sufficient and relevant information to answer the question.\n\nInstructions:\n1. Internally, perform a detailed step-by-step analysis (chain-of-thought) of the clinical question and the clinical note.\n2. Decide whether the clinical note provides a clear answer to the question.\n3. Return only \"Yes\" if the note contains sufficient evidence to answer the question, or \"No\" otherwise.\n\nOutput Format:\nA single word: Yes or No\n\nReminder: Use your internal chain-of-thought to reason through the task, but do not include any explanation or reasoning in the output. Only return Yes or No.",
    "user": "Patient Narrative:\n {patient_narrative}\n\nClinical Question:\n {clinical_question} \n\nClinical Note:\n {sentences}"
}
\end{lstlisting}

\subsubsection{CoT + Few-Shot Prompt}
\begin{lstlisting}[language=json, breaklines=true]
{
    "system": "Role:\nYou are a medical expert with advanced expertise in clinical natural language processing (NLP). You specialize in analyzing unstructured clinical notes to extract medically relevant information and evaluate patient narratives. Your clinical acumen allows you to understand complex medical language and determine whether a given note contains sufficient evidence to answer specific clinical questions.\n\nTask:\nGiven a clinical note and a clinical question, determine whether the clinical note contains enough explicit and relevant information to confidently answer the question.\n\nInstructions:\n1. Internally, conduct a detailed chain-of-thought analysis to interpret the clinical question and assess the content of the note.\n2. Judge whether the clinical note includes clear, sufficient, and directly relevant information that supports and answers the question. Ignore any notes or phrases that are non-informative or purely structural, such as headers (e.g., \"Brief Hospital Course:\") or general section labels without medical content.\n3. Return only one word based on your internal reasoning:\n   - \"Yes\" — if the note contains clear evidence to answer the question.\n   - \"No\" — if the note lacks sufficient or relevant evidence to confidently answer the question.\n\nOutput Format:\nRespond with a single word only: Yes or No. Do not include any reasoning or explanation in your response.\n\nExample:\n\nPatient Narrative:\nTook my 59 yo father to ER ultrasound discovered he had an aortic aneurysm. He had a salvage repair (tube graft). Long surgery / recovery for couple hours then removed packs. Why did they do this surgery? After this time he spent 1 month in hospital now sent home.\n\nClinical Question:\nWhy did they perform the emergency salvage repair on him?\n\nClinical Notes & Answers:\nHe was transferred to the hospital on 2025-1-20 for emergent repair of his ruptured thoracoabdominal aortic aneurysm.\nAnswer: Yes\n\nHe was immediately taken to the operating room where he underwent an emergent salvage repair of ruptured thoracoabdominal aortic aneurysm with a 34-mm Dacron tube graft using deep hypothermic circulatory arrest.\nAnswer: Yes\n\nPlease see operative note for details which included cardiac arrest x2.\nAnswer: No\n\nPostoperatively he was taken to the intensive care unit for monitoring with an open chest.\nAnswer: No\n\nHe remained intubated and sedated on pressors and inotropes.\nAnswer: No\n\nOn 2025-1-22, he returned to the operating room where he underwent exploration and chest closure.\nAnswer: No\n\nOn 1-25 he returned to the OR for abd closure JP/ drain placement/ feeding jejunostomy placed at that time for nutritional support.\nAnswer: No\n\nThoracoabdominal wound healing well with exception of very small open area mid wound that is @1cm around and 1/2cm deep, no surrounding erythema.\nAnswer: Yes\n\nPacked with dry gauze and covered w/DSD.\nAnswer: No",
    "user": "Patient Narrative:\n {patient_narrative}\n\nClinical Question:\n {clinical_question} \n\nClinical Note:\n {sentences}"
}
\end{lstlisting}

\subsection{Prompt for Second Step Argumentation Creation}
\begin{lstlisting}[language=json, breaklines=true]
{
    "system":"You are a medical expert specializing in clinical natural language processing. Your task is to generate an answer to a patient’s health-related question based only on the information provided in the clinical notes. Write a short, medically sound answer that either paraphrases or argues or summaries using the key phrases from the notes. The response must not exceed 75 words. Focus only on the clinical notes provided. Your output must be a single, focused paragraph of 75 words or fewer — never exceed this limit. Give only the answer, without any additional information or explanations.",
    "user": "Patient Narrative:\n {patient_narrative}\n\nClinical Question:\n {clinical_question} \n\nClinical Notes:\n {sentences}",
    "assistant": ""
}
\end{lstlisting}


\end{document}